\documentclass{article}
\usepackage{PRIMEarxiv}
\usepackage{xcolor}
\usepackage{pifont}
\usepackage{tabularx}
\usepackage{array}
\usepackage{amsmath, amsfonts, amsthm, amssymb, epsfig}
\usepackage{authblk}
\usepackage[utf8]{inputenc}
\usepackage[T1]{fontenc}
\usepackage{hyperref}
\usepackage{url}
\usepackage{booktabs}
\usepackage{amsfonts}
\usepackage{nicefrac}
\usepackage{microtype}
\usepackage{lipsum}
\usepackage{fancyhdr}
\usepackage{graphicx}
\usepackage{subcaption}
\usepackage{algorithm}
\usepackage{algpseudocode}
\usepackage[absolute,overlay]{textpos}
\usepackage{makecell}

\title{GPEC: Efficient Pre-LLM Gaussian Process Embedding Correction for Cardiac Video Caption Generation}
\author{
Arefeh Rezaei\\
Faculty of Computer Engineering, K.N. Toosi University of Technology,\\
Tehran, Iran\\
\href{mailto:a_rezai@alumni.kntu.ac.ir}{a\_rezai@alumni.kntu.ac.ir}
}

\begin{document}

\maketitle

\begin{abstract}

Multimodal large language models (MLLMs) have shown promising capabilities for video understanding and caption generation, but their performance can degrade when applied to specialized medical imaging domains such as echocardiography. This work presents Gaussian Process Embedding Correction (GPEC), a modular and computationally efficient pre-LLM error-correction method designed to improve the visual representations used by VideoChat2 for cardiac ultrasound caption generation. GPEC is inserted between the visual projection stage and the language model and learns to predict a residual correction that moves the projected visual representation toward an annotation-guided target representation. The target is constructed by transforming structured video annotations into qualitative attributes, instantiating a fixed-format reference caption, and mapping the resulting caption into the language-model embedding space. The correction is modeled using a sparse variational Gaussian Process with inducing points, natural-parameter variational updates, and a block-wise linear kernel formulation, while the original VideoChat2 components remain frozen during training and inference.The proposed method is evaluated using complementary representation-level, caption-level, content-oriented, and execution-time measures. The experiments compare the original VideoChat2 model with the proposed VideoChat2 + GPEC configuration under the same input and reference conditions. The results show improvements in caption similarity and content alignment after applying the proposed pre-LLM correction. Moreover, the addition of GPEC introduces less than 0.05~s of inference-time overhead per video under the evaluated experimental setting, demonstrating that the proposed correction can improve caption generation with a small additional computational cost. These findings support GPEC as a modular and computationally efficient approach for improving caption generation in specialized medical video domains without requiring end-to-end fine-tuning of the pretrained multimodal backbone.Code: \href{https://github.com/areferezaee/GPCE}{https://github.com/areferezaee/GPCE}

\end{abstract}



\section{Introduction}
\label{introduction}

Recent advances in vision-language models have enabled large language models (LLMs) to process visual information and generate natural-language descriptions of complex visual content~\cite{tang2025video, maaz2023videochatgpt}. Extending these capabilities to video introduces the additional challenge of modeling temporal visual information, leading to the development of video-language models for video understanding and description generation~\cite{lin2024videollava, xu2024pllava, zhang2023videollama, li2023llamavid}. Despite these advances, transferring the visual and semantic information required for accurate language generation remains challenging in specialized domains, where domain-specific data, terminology, and clinical context can limit the generalization of vision-language models~\cite{ryu2025medicalvlm}.

This challenge is particularly relevant to cardiac video understanding\cite{huang2026echomllm}. Echocardiography provides rich video-based information about cardiac structure and function and has motivated the development of dedicated vision-language models for echocardiographic interpretation~\cite{christensen2024echoclip}. More recently, video-based vision-language foundation models have been developed specifically for comprehensive echocardiographic analysis, further demonstrating the growing role of multimodal learning in this domain~\cite{vukadinovic2026echoprime}. At the same time, individual examinations may differ in functional characteristics and other annotation-derived attributes that are important for generating sample-specific descriptions~\cite{ouyang2020video}. A model may therefore capture the general visual characteristics of cardiac ultrasound while still producing descriptions that are incorrectly grounded, overly generic, or incomplete with respect to the characteristics of the individual sample~\cite{liu2026heartbeat, huang2026echomllm}.

In preliminary experiments, the pretrained VideoChat2 model~\cite{li2024mvbench} was observed to produce captions that could be semantically inconsistent with the underlying cardiac content. For example, cardiac ultrasound videos could be described as non-cardiac ultrasound studies or receive generic descriptions unrelated to the specific ventricular motion represented in the input. Importantly, these observations were obtained without modifying or fine-tuning the original VideoChat2 model. This motivated an investigation into whether part of the discrepancy between the visual input and the generated language could be addressed before the language-generation stage, directly at the interface between the visual representation and the language model.

To address this issue, this work introduces GPEC (Gaussian Process Embedding Correction), a pre-LLM embedding correction framework for cardiac video caption generation. Rather than attempting to correct an erroneous caption after decoding, GPEC operates directly on the representation provided to the language model. The proposed module learns a residual correction that moves the projected visual representation toward a target representation derived from structured, annotation-based reference captions. The corrected representation is then passed to the language model to generate the final caption.

The target construction is based on a fixed caption template. For each cardiac video, the available annotations are transformed into qualitative attributes and inserted into the template, producing a standardized reference caption while preserving the characteristics that distinguish one sample from another. The resulting caption is subsequently mapped through the language model's own input embedding space, providing a target representation in the same latent space as the projected visual features. This formulation directly connects the annotation-derived information to the representation being corrected, without introducing a separate feature space for supervision.

A central challenge of the proposed setting arises from the high dimensionality of the visual representation compared with the relatively limited number of samples used to train the correction module. Although the underlying dataset provides a substantial collection of cardiac videos, the GPEC module learns the correction from a comparatively small set of annotated training examples relative to the dimensionality of the representation. This results in a challenging high-dimensional learning problem, a setting that is also recognized as difficult in echocardiography-based learning~\cite{nilsson2024gvit}. GPEC addresses this challenge using a sparse variational Gaussian Process with inducing points, natural-parameter variational updates, and a block-wise linear kernel construction. Despite using the limited number of training samples for the correction module, the resulting representation correction leads to improved caption generation on the many evaluated test samples.

An important characteristic of GPEC is that it does not require end-to-end fine-tuning of the pretrained multimodal model. The visual encoder, visual projection module, and language model remain frozen, while the proposed correction module is trained separately and inserted between the visual projection stage and the language model. This design makes GPEC modular and allows the effect of pre-LLM representation correction to be evaluated independently of changes to the underlying video-language backbone. In addition, the proposed module introduces only a small measured inference overhead under the evaluated experimental setting.

Overall, GPEC addresses the mismatch between visual representations and language generation at the representation level rather than attempting to repair the generated text after decoding. The approach is designed for a challenging specialized-video setting in which cardiac videos share substantial visual structure while exhibiting sample-specific functional differences. By combining annotation-guided target construction with probabilistic representation correction, GPEC aims to improve the grounding and quality of captions generated by a frozen video-language model.

The main contributions of this work are:

\begin{enumerate}

\item A pre-LLM Gaussian Process Embedding Correction (GPEC) framework for improving cardiac video caption generation without fine-tuning the original VideoChat2 model.

\item An annotation-guided target construction strategy that converts structured cardiac annotations into qualitative, fixed-format reference captions and maps them into the language-model embedding space.

\item A sparse variational Gaussian Process formulation with inducing points and natural-parameter updates for learning high-dimensional representation corrections.

\item A block-wise linear kernel construction designed for the high-dimensional visual representation used by VideoChat2.

\item An experimental evaluation of representation-level correction, caption-generation quality, content-oriented caption alignment, and inference-time overhead.

\end{enumerate}

\section{Background and Preliminaries}
\label{background}

\subsection{EchoNet-Dynamic}

EchoNet-Dynamic is a publicly available echocardiography dataset introduced for the development and evaluation of video-based methods for cardiac function assessment~\cite{ouyang2020video}. The dataset contains 10,030 apical four-chamber echocardiography videos acquired at Stanford University Hospital and provides expert-derived cardiac measurements and annotations, including left ventricular ejection fraction (LVEF), end-diastolic volume (EDV), and end-systolic volume (ESV), together with information related to cardiac-cycle phases and ventricular boundaries~\cite{ouyang2020video}. The videos capture cardiac motion across multiple cardiac cycles, making the dataset suitable for studying the relationship between temporal echocardiographic appearance and cardiac functional characteristics.

The dataset was prepared using standardized preprocessing of the echocardiographic videos, including cropping of the relevant imaging region and normalization of the spatial representation. In addition to the video data, the associated annotations provide quantitative measurements and cardiac-phase information that can be used to characterize ventricular motion and function~\cite{ouyang2020video}.

In this work, EchoNet-Dynamic provides both the cardiac video inputs and the annotation information used to construct the supervision for GPEC. The annotation values associated with the selected videos are transformed into qualitative attributes and incorporated into a fixed-format reference caption. These annotation-derived descriptions are subsequently mapped into the language-model embedding space to construct the target representation used for training the proposed representation-correction module.

\subsection{VideoChat2}

VideoChat2 is a multimodal large language model designed for video understanding and video-language interaction. It was introduced together with MVBench, a comprehensive benchmark for evaluating temporal and multimodal reasoning in video-language models. The model is developed through progressive multimodal training, in which visual information is aligned with a large language model and subsequently refined using multimodal instruction-tuning data~\cite{li2024mvbench}.

VideoChat2 connects video-derived visual representations to a language model through a multimodal alignment mechanism. Its architecture incorporates a visual encoder followed by a Q-Former that compresses visual information into a smaller set of query representations, which are subsequently projected and provided to the language model. The model is developed through a progressive multimodal training strategy, including vision-language alignment, visual-language connection to the LLM, and instruction tuning, enabling the integration of visual and linguistic information for video-understanding tasks~\cite{li2024mvbench}.

In this work, VideoChat2 is used as the pretrained video-language backbone for cardiac echocardiography caption generation. Given an echocardiographic video, the visual pipeline produces projected visual embeddings after the Q-Former and projection stages. In the configuration used in this work, the projected representation consists of 96 visual tokens with a hidden dimension of 3072. This projected representation is provided to the proposed GPEC module before being passed to the language model.

The pretrained VideoChat2 parameters remain frozen throughout the proposed framework. In particular, GPEC does not modify the visual encoder, Q-Former, visual projection module, or language-model parameters. Instead, it operates on the projected visual representation produced by the frozen backbone and applies a learned correction before the representation is passed to the language model for caption generation.

\subsection{Gaussian Process Regression}

Gaussian process (GP) regression provides a non-parametric Bayesian framework for modeling unknown functions by placing a probability distribution over possible functions. A Gaussian process is characterized by a mean function and a covariance function (kernel), and is written as

$$
f(x) \sim \mathcal{GP}\left(m(x),k(x,x')\right),
$$

where $m(x)$ denotes the mean function and $k(x,x')$ defines the covariance between function values evaluated at $x$ and $x'$~\cite{rasmussen2006gaussian}.

Given a set of observations, the covariance function determines how information is shared between different inputs. In the case of a Gaussian likelihood, the posterior distribution remains Gaussian and its mean and covariance can be obtained analytically. For a set of training inputs $X$ and corresponding observations $\mathbf{y}$, let $K_{XX}$ denote the kernel matrix and let $K_{*X}$ denote the covariance between a test input $x_*$ and the training inputs. Under a zero-mean function assumption, the posterior mean at $x_*$ is given by

$$
\mu(x_*) = K_{*X}K_{XX}^{-1}\mathbf{y},
$$

while the posterior covariance is obtained by conditioning the prior Gaussian process on the observed data~\cite{rasmussen2006gaussian}.

The kernel function is a central component of GP regression because it determines the similarity structure imposed on the input space. Different kernel functions encode different assumptions about how function values vary with their inputs. In this work, the GP is used to model the relationship between the projected visual representation and its corresponding embedding correction residual. The specific sparse variational formulation and block-wise kernel construction used for this purpose are described in the following sections.

\subsection{Variational Gaussian Processes}

Exact Gaussian Process inference requires operations on covariance matrices whose computational and memory costs increase with the number of training observations. Sparse variational formulations alleviate this limitation by introducing a smaller set of inducing variables that provide a compact representation of the latent function and enable scalable approximate inference~\cite{hensman2015scalable}.

Let $u$ denote the latent function values evaluated at $M$ inducing points. Instead of explicitly representing the latent function at all training inputs, a variational distribution is introduced over the inducing variables:

$$
q(u)=\mathcal{N}(u\mid\mu,\Sigma).
$$

The inducing variables provide a connection between the latent function and the observed inputs through the corresponding covariance matrices. In particular, if $K_{mm}$ denotes the covariance matrix between inducing points and $K_{nm}$ denotes the covariance between observed inputs and inducing points, the conditional relationship between the inducing variables and function values at the observed inputs involves

$$
K_{nm}K_{mm}^{-1}.
$$

Variational inference then optimizes a lower bound on the marginal log likelihood, commonly referred to as the evidence lower bound (ELBO):

$$
\mathcal{L}_{\mathrm{ELBO}}
=
\mathbb{E}_{q(f)}
\left[
\log p(y\mid f)
\right]
-
\mathrm{KL}
\left[
q(u)\,\|\,p(u)
\right].
$$

The first term measures the expected agreement between the latent function and the observations under the variational posterior, while the KL divergence regularizes the variational distribution toward the GP prior~\cite{hensman2015scalable}.

The sparse variational formulation therefore provides a practical framework for approximate posterior inference using a limited number of inducing variables rather than explicitly optimizing the full latent covariance structure. In the proposed framework, this formulation is used to model the residual between the projected visual representation and the annotation-derived target representation. The specific inducing-point parameterization, natural-parameter variational updates, and block-wise kernel construction adopted for GPEC are described in Section~\ref{method}.

\section{Method}
\label{method}

This work introduces GPEC (Gaussian Process Embedding Correction), an independent pre-LLM representation-correction module inserted between the visual projection stage of VideoChat2 and its language model. The objective of GPEC is to learn a representation-level correction that reduces the discrepancy between the projected visual representation and an annotation-guided target representation before language generation.

The pretrained VideoChat2 model remains frozen throughout the proposed procedure. In particular, the visual encoder, Q-Former, visual projection module, and language model are not fine-tuned. Given an input cardiac video $V_i$, VideoChat2 first produces a projected visual representation $Z_i^p$. GPEC then predicts a residual correction from this representation. The predicted correction is added to the original projection, and the resulting corrected representation is passed to the frozen language model for caption generation.

The overall inference process can be summarized as

$$
V_i
\rightarrow
\mathrm{VideoChat2}
\rightarrow
Z_i^p
\rightarrow
\mathrm{GPEC}
\rightarrow
\widehat{Z}_i^c
\rightarrow
\mathrm{LLM}
\rightarrow
\widehat{C}_i.
$$

Thus, GPEC operates entirely in the representation space between visual projection and language generation without modifying the pretrained multimodal backbone.

\subsection{Annotation-Guided Target Construction}

GPEC is trained using target representations derived from the available annotations associated with each echocardiography video. Rather than directly inserting the original numerical annotation values into the supervision signal, the annotations are deterministically converted into qualitative descriptors and then used to construct a standardized reference caption. This procedure preserves annotation-specific information while maintaining a common linguistic structure across samples.

For each video $V_i$, the available annotations include left ventricular ejection fraction (EF), end-diastolic volume (EDV), and end-systolic volume (ESV). These numerical measurements are mapped to qualitative attributes through predefined classification rules. For an annotation vector

$$
\mathcal{A}_i=(EF_i,EDV_i,ESV_i),
$$

the corresponding qualitative descriptors are obtained as

$$
\mathcal{Q}_i=\mathcal{C}(\mathcal{A}_i),
$$

where $\mathcal{C}(\cdot)$ denotes the deterministic classification procedure.

The EF value determines four qualitative characteristics: left ventricular systolic function, ventricular contraction, ventricular emptying, and overall pumping performance. EF values of 55 or higher are mapped to \textit{preserved}, \textit{marked}, \textit{effective}, and \textit{good}, respectively. EF values from 40 to below 55 are mapped to \textit{mildly reduced}, \textit{moderate}, \textit{moderately effective}, and \textit{moderate}, respectively. Values below 40 are mapped to \textit{reduced}, \textit{limited}, \textit{limited}, and \textit{reduced}.

EDV is used to characterize ventricular filling relative to the empirical EDV distribution of the dataset. Values below the 33rd percentile are mapped to \textit{relatively limited} filling, values from the 33rd to below the 67th percentile to \textit{moderate} filling, and values at or above the 67th percentile to \textit{substantial} filling. The post-contraction residual cavity is characterized using the normalized ratio $ESV/EDV$. Ratios below 0.35 are classified as \textit{small}, ratios from 0.35 to below 0.60 as \textit{moderate}, and ratios of 0.60 or greater as \textit{substantial}.

The resulting qualitative descriptors are inserted into the following fixed-format caption:
\begin{quote}
\textit{The video shows the cardiac motion of the heart throughout the cardiac cycle that should be described accurately. The left ventricle shows [filling] filling during diastole, followed by a [contraction] reduction in ventricular cavity size during systole. The cavity becomes [residual] after contraction, indicating [emptying] ventricular emptying. Overall, the observed ventricular motion is consistent with [systolic function] left ventricular systolic function and [pumping] pumping performance.}
\end{quote}

The bracketed fields are populated with the qualitative descriptors obtained from the corresponding annotations. Consequently, all reference captions share the same linguistic structure while retaining sample-specific information. The resulting caption is denoted by $C_i^{*}$ and serves both as the annotation-derived reference description and as the source for constructing the target representation.

The reference caption is subsequently tokenized using the tokenizer associated with the same language model used by VideoChat2 and mapped through the frozen input embedding layer:

$$
Z_i^{t}
=
E_{\mathrm{LLM}}
\left(
\mathrm{Tokenizer}(C_i^{*})
\right),
$$

where $E_{\mathrm{LLM}}$ denotes the language-model input embedding function.

The projected visual representation and target representation are defined in the same latent space. In the configuration used in this work,

$$
Z_i^p,Z_i^t\in\mathbb{R}^{96\times3072}.
$$

In the implementation, the first 96 caption-embedding positions are retained so that the target representation matches the shape of the projected visual representation. This target is then used for residual learning.

\subsection{Residual Embedding Correction}

Rather than directly predicting the target representation, GPEC formulates the correction problem as residual learning. For each training sample, the desired correction is defined as

$$
\Delta Z_i=Z_i^t-Z_i^p,
$$

where $Z_i^p$ is the original projected visual representation and $Z_i^t$ is the annotation-derived target representation.

The residual therefore has the same shape as the projected representation:

$$
\Delta Z_i\in\mathbb{R}^{96\times3072},
$$

corresponding to 294,912 representation coordinates after flattening.

GPEC models the probabilistic mapping from the projected representation to the residual:

$$
f_{\mathrm{GPEC}}:Z_i^p\rightarrow\Delta Z_i.
$$

At inference time, the target representation is unavailable. Instead, the learned model produces a predictive residual from the projected representation:

$$
\widehat{\Delta Z}_i
=
f_{\mathrm{GPEC}}(Z_i^p),
$$

and the corrected representation is obtained as

$$
\widehat{Z}_i^c
=
Z_i^p+\widehat{\Delta Z}_i.
$$

The corrected representation is then passed to the frozen language model to generate the final caption. This formulation separates representation correction from language generation: GPEC estimates the discrepancy between the projected visual representation and the annotation-guided target, while the language model remains responsible for decoding the corrected representation into natural language.

\subsection{Sparse Variational Gaussian Process}

The residual prediction problem is modeled using a sparse variational Gaussian Process. The projected visual representation serves as the GP input, while the residual representation is used as the regression target.

The projected representation contains 96 tokens with a hidden dimension of 3072 and is flattened before being processed by the GP:

$$
D=96\times3072=294{,}912.
$$

Let $M$ denote the number of inducing points. In the implementation, $M=11$. Let $K_{mm}\in\mathbb{R}^{M\times M}$ denote the covariance matrix among the inducing points, $K_{nm}\in\mathbb{R}^{N\times M}$ the covariance between observed training inputs and inducing points, and $K_{nn}$ the covariance matrix among observed inputs. The inducing-to-observation covariance mapping is defined as

$$
\kappa=K_{nm}K_{mm}^{-1}.
$$

In practice, $K_{mm}^{-1}$ is not explicitly formed. Instead, a numerically stabilized Cholesky factorization of $K_{mm}$ is used together with triangular solves to compute the required linear-system solution.

The variational posterior over the inducing variables is

$$
q(u)=\mathcal{N}(u\mid\mu,\Sigma).
$$

The implementation parameterizes this posterior using natural parameters:

$$
\eta=\Sigma^{-1}\mu,
$$

and

$$
H=-\frac{1}{2}\Sigma^{-1}.
$$

The covariance and mean can therefore be recovered as

$$
\Sigma=(-2H)^{-1},
$$

and

$$
\mu=\Sigma\eta.
$$

A shared covariance $\Sigma$ is maintained across the output dimensions, whereas the natural parameter $\eta$ is maintained separately for the output dimensions of the residual representation.

Given the variational parameters, the predictive mean at the observed locations is

$$
\mu_f=\kappa\mu.
$$

The latent predictive variance is computed as

$$
\mathrm{Var}[f]
=
\operatorname{diag}
\left(
K_{nn}-K_{nm}K_{mm}^{-1}K_{mn}
\right)
+
\operatorname{diag}
\left(
\kappa\Sigma\kappa^{\top}
\right).
$$

In the implementation, the resulting pointwise predictive variance is expanded across the output dimensions. Thus, the predictive uncertainty is shared across output coordinates at a given input, while the predictive mean remains output-specific.

The residual targets are modeled with a Gaussian likelihood having fixed observation noise variance

$$
\sigma^2=10^{-4}.
$$

The variational objective is the evidence lower bound:

$$
\mathcal{L}_{\mathrm{ELBO}}
=
\mathbb{E}_{q(f)}
\left[
\log p(Y\mid f)
\right]
-
\mathrm{KL}
\left[
q(u)\,\|\,p(u)
\right].
$$

The KL term depends on the inducing-point prior covariance $K_{mm}$ and the variational parameters $\mu$ and $\Sigma$. This sparse variational formulation provides a tractable probabilistic model for the high-dimensional residual without explicitly maintaining a full covariance matrix over all representation coordinates.

\subsection{Natural-Parameter Variational Update}

The variational posterior is updated directly in natural-parameter space. Consider a minibatch containing $N_{\mathrm{batch}}$ observations from a training set of size $N$. The minibatch contribution to the likelihood is scaled by

$$
s=\frac{N}{N_{\mathrm{batch}}}.
$$

Let $Y$ denote the residual targets associated with the minibatch. Under the Gaussian likelihood, the target natural parameter associated with the mean is

$$
\eta_{\mathrm{target}}
=
\frac{s}{\sigma^2}
\kappa^{\top}Y,
$$

while the second natural parameter is

$$
H_{\mathrm{target}}
=
-\frac{1}{2}
\left(
K_{mm}^{-1}
+
\frac{s}{\sigma^2}
\kappa^{\top}\kappa
\right).
$$

The first term in $H_{\mathrm{target}}$ corresponds to the inducing-point prior precision, whereas the second term represents the precision contribution induced by the minibatch observations under the Gaussian likelihood.

Rather than replacing the current variational parameters directly with their target values, damped natural-parameter updates are applied:

$$
\eta
\leftarrow
\eta+
\rho
\left(
\eta_{\mathrm{target}}-\eta
\right),
$$

and

$$
H
\leftarrow
H+
\rho
\left(
H_{\mathrm{target}}-H
\right),
$$

where $\rho$ denotes the natural-parameter update rate.

Following each update, the variational covariance and mean are recovered using

$$
\Sigma=(-2H)^{-1},
\qquad
\mu=\Sigma\eta.
$$

The required matrix operations are implemented using Cholesky-based linear solves rather than explicit matrix inversion whenever applicable. This update scheme allows the variational posterior to be adapted using minibatches while retaining output-specific mean-related natural parameters for the high-dimensional residual.

\subsection{Block-Wise Kernel Construction}

To construct the covariance function for the high-dimensional projected representation, the implementation divides the input feature dimension into 12 contiguous blocks of equal dimensionality. Because

$$
294{,}912/12=24{,}576,
$$

each block contains 24,576 input features. Let the block representation be written as

$$
x=
[x^{(1)},x^{(2)},\ldots,x^{(12)}].
$$

A separate linear kernel is applied to each block, and the resulting covariance matrices are summed:

$$
k(x,x')
=
\sum_{r=1}^{12}
k_r
\left(
x^{(r)},x'^{(r)}
\right).
$$

Each block kernel is implemented as a linear kernel wrapped by a scale kernel. The base variance of the linear kernel is set to $10^{-7}$. A small diagonal jitter term is added to each block covariance before the covariance matrices are aggregated, providing numerical stabilization for subsequent Cholesky-based computations.

This construction decomposes the kernel evaluation across contiguous subsets of the input feature dimension while retaining a single aggregated covariance matrix for the GP. The block decomposition applies to the GP input features rather than to the output residual dimensions. The residual representation remains jointly modeled through the shared variational covariance and output-specific natural mean parameters.

\subsection{Predictive Correction and Inference}

After training, the learned variational posterior is used to predict the residual representation for an unseen cardiac video. Given a projected representation $Z^p$, the corresponding kernel matrices are computed and the inducing-to-input mapping is obtained as

$$
\kappa=K_{nm}K_{mm}^{-1}.
$$

The predictive mean of the residual is then computed as

$$
\widehat{\Delta Z}
=
\kappa\mu.
$$

The corrected representation is constructed by

$$
\widehat{Z}^{c}
=
Z^{p}
+
\widehat{\Delta Z}.
$$

The predictive mean is used directly for the representation correction. Although the GP predictive variance is available from the variational posterior, it is not used to modify the representation during inference.

The corrected representation is then supplied to the frozen language model:

$$
V
\rightarrow
Z^{p}
\rightarrow
\widehat{\Delta Z}
\rightarrow
\widehat{Z}^{c}
\rightarrow
\mathrm{LLM}
\rightarrow
\widehat{C}.
$$

No annotation, reference caption, or target embedding is required during inference. Annotation-derived targets are used only during training to learn the residual correction function.

Overall, GPEC introduces a modular representation-correction stage that operates independently of the pretrained VideoChat2 parameters and learns to reduce the discrepancy between the projected visual representation and an annotation-guided target before language generation.

\section{Experiments and Results}
\label{experiments}

\subsection{Experimental Setup}

The proposed Gaussian Process Embedding Correction (GPEC) module is evaluated for improving caption generation by VideoChat2 on cardiac echocardiography videos. The reported caption-level evaluation is conducted on 25 retained test samples, together with their corresponding annotation information and annotation-derived reference captions.

For each video, the associated annotations are converted into a standardized reference caption using the fixed caption-generation procedure described in Section~\ref{method}. The numerical annotation values are first mapped to qualitative descriptors and then inserted into a fixed linguistic template. Consequently, all reference captions share the same overall semantic structure while preserving the sample-specific characteristics derived from the annotations. The same annotation-derived captions are used as reference descriptions for both the baseline and the proposed configuration.

The original pretrained VideoChat2 model serves as the baseline. In the proposed configuration, GPEC is inserted between the visual projection stage and the language model. The visual encoder, Q-Former, visual projection module, and language model remain frozen, while GPEC predicts a residual correction to the projected visual representation. Both configurations receive the same cardiac video inputs and are evaluated against the same corresponding reference captions.

For each test video, the baseline generates a caption directly from the original projected representation. In the proposed configuration, the projected representation is first corrected by GPEC, after which the corrected representation is passed to the frozen language model to generate the final caption. This paired evaluation isolates the effect of pre-LLM representation correction on the generated captions.

\subsection{Implementation Details}

The projected visual representation produced by VideoChat2 consists of 96 visual tokens with a hidden dimension of 3072, resulting in a representation of size $96\times3072$ and a flattened dimensionality of 294,912.

For each video, the target representation is constructed from its annotation-derived reference caption using the procedure described in Section~\ref{method}. The caption is tokenized using the tokenizer associated with the VideoChat2 language model and mapped through the frozen language-model input embedding layer. The first 96 embedding positions are retained so that the target and projected visual representations have the same shape:

$$
Z_i^p,Z_i^t\in\mathbb{R}^{96\times3072}.
$$

GPEC is implemented as a sparse variational Gaussian Process with 11 inducing points. The variational posterior is parameterized using natural parameters. The GP input representation is divided into 12 contiguous blocks of equal dimensionality. Since the flattened representation has 294,912 features, each block contains 24,576 features. A linear kernel wrapped by a scale kernel is applied independently to each block, and the resulting covariance matrices are summed to form the overall covariance. The base variance of each linear kernel is set to $10^{-7}$, and a small diagonal jitter term is added for numerical stabilization.

A fixed Gaussian observation noise variance of $10^{-4}$ is used in the variational objective. The variational natural parameters are updated using damped natural-parameter updates. During inference, the predictive mean of the Gaussian Process is used as the estimated embedding residual. The corrected representation is computed as

$$
Z_i^c
=
Z_i^p+\widehat{\Delta Z}_i,
$$

and is subsequently passed to the frozen VideoChat2 language model for caption generation.

All experiments use the same pretrained VideoChat2 backbone, with its visual encoder, Q-Former, visual projection module, and language model kept frozen. Only the GPEC correction module is trained.

\subsection{Evaluation Metrics}

The proposed approach is evaluated at three complementary levels: representation-level correction, caption-generation quality, and content-oriented caption analysis.

\paragraph{Representation-level metrics.}

The learned embedding correction is evaluated using mean squared error (MSE), mean absolute error (MAE), and negative log predictive density (NLPD). MSE and MAE quantify the discrepancy between the predicted residual representation and the corresponding target representation, while NLPD evaluates the probabilistic quality of the Gaussian Process prediction by taking predictive uncertainty into account. These metrics are obtained directly from the GPEC model evaluation and are therefore reported separately from the caption-generation metrics.

\paragraph{Caption-level metrics.}

Generated captions are compared with the annotation-derived reference captions using BLEU-1, BLEU-2, BLEU-3, and BLEU-4, which quantify lexical overlap at different $n$-gram orders. ROUGE-L provides a complementary sequence-level similarity measure based on the longest common subsequence, while METEOR evaluates word-level similarity using flexible matching between the generated and reference captions. CIDEr is additionally reported using TF-IDF-weighted $n$-gram similarity for caption evaluation.

\paragraph{Content-oriented metrics.}

Lexical similarity alone does not necessarily indicate whether the information encoded in the reference description is preserved. Therefore, a complementary content-oriented evaluation is performed using the recurring information fields represented in the annotation-derived captions. These fields include cardiac structure, cardiac-cycle phase, diastolic filling, systolic cavity reduction, post-contraction cavity state, ventricular emptying, left ventricular systolic function, and pumping performance.

Three complementary measures are used. Clinical Field Recall measures the proportion of relevant reference fields that are identified in the generated caption. Exact Clinical Attribute Accuracy measures whether the qualitative attribute associated with a field is reproduced exactly. Clinical Attribute Agreement provides a softer measure when the predicted attribute is not identical to the reference but is considered partially or closely matched according to the field-specific agreement rules. These measures are used as complementary content-oriented caption measures and are not intended to represent clinical diagnosis, clinical decision-making, or medical performance.

The content-oriented evaluation is implemented using a fixed, rule-based ontology derived from the structure of the annotation-based reference captions. For each predefined field, a set of lexical patterns is specified to determine whether the corresponding concept is present in the reference or generated caption. Regular-expression matching is applied after text normalization, including lowercasing and whitespace normalization. The same predefined field definitions are applied consistently to both reference and generated captions.

For fields with qualitative attributes, additional lexical patterns are defined for the possible attribute values. The first matching attribute pattern is assigned as the extracted value for that field. Thus, the evaluation does not compare entire captions as exact strings; instead, it decomposes each caption into a set of predefined content fields and their associated attributes and then compares the extracted information between the reference and generated caption.

Clinical Field Recall is computed only over fields that are present in the reference caption. For each such field, a value of one is assigned when the corresponding field is detected in the generated caption and zero otherwise. The field-level recall is then calculated as the mean of these binary field-detection scores.

For fields for which a reference attribute value can be extracted, Exact Clinical Attribute Accuracy is computed by comparing the extracted prediction value with the corresponding reference value. An exact match receives a score of one, whereas a mismatch or a missing predicted attribute receives a score of zero.

Clinical Attribute Agreement provides a graded comparison for selected ordered attributes. For these fields, the possible attribute values are assigned an ordinal severity or functional ordering, and the agreement score decreases as the distance between the predicted and reference values increases. For example, attributes such as diastolic filling, ventricular emptying, left ventricular systolic function, and pumping performance are evaluated using field-specific ordered scales. Consequently, a prediction that is not identical to the reference can still receive partial agreement when it corresponds to a neighboring or intermediate qualitative level.

This content-oriented evaluation is therefore a structured lexical matching procedure based on a predefined ontology and field-specific rules. It is intended to quantify the preservation of predefined information contained in the reference captions and should not be interpreted as a general-purpose semantic understanding measure or as a clinical diagnostic evaluation.

\subsection{Efficiency Analysis}

To assess the practical efficiency of the proposed pre-LLM correction strategy, I additionally measure the execution cost of the baseline and corrected configurations under the same hardware and inference conditions. The additional runtime introduced by GPEC is measured across the evaluated test videos and is summarized as an average overhead of $0.028 \pm 0.01$~s per video.

Because the original VideoChat2 parameters remain frozen, the efficiency analysis focuses on the computational overhead introduced by the additional correction module rather than the cost of retraining or fine-tuning the underlying VideoChat2 model. The reported overhead corresponds specifically to the additional execution time required by GPEC during inference under otherwise identical conditions.

\subsection{Baseline Models}

The original pretrained VideoChat2 model serves as the primary baseline. It generates captions directly from the projected visual representation without any embedding correction.

The proposed configuration, denoted as VideoChat2 + GPEC, uses the same pretrained VideoChat2 model while introducing the GPEC module between the visual projection stage and the language model. Since the parameters of the original VideoChat2 components remain frozen, the comparison isolates the effect of the proposed pre-LLM embedding correction mechanism.

\subsection{Quantitative Results}

The quantitative evaluation compares VideoChat2 and VideoChat2 + GPEC at the representation, caption, and content levels. The final caption-level and content-oriented results are computed over 25 test samples.

\subsubsection{Embedding-Level Results}

Table~\ref{tab:embedding_results} reports the embedding-level performance of GPEC. These metrics are obtained from the direct evaluation of the trained Gaussian Process model.

\begin{table}[!th]
\centering
\caption{Embedding-level evaluation of the GPEC correction on the test set. Lower values indicate better performance.}
\label{tab:embedding_results}
\begin{tabular}{lccc}
\midrule
Model & MSE $\downarrow$ & MAE $\downarrow$ & NLPD $\downarrow$ \\
\midrule
GPEC & 0.0059 & 0.0508 & -0.4432 \\
\midrule
\end{tabular}
\end{table}

\subsubsection{Caption-Level Results}

Table~\ref{tab:caption_results} summarizes the caption-generation performance over the 25 test samples.

\begin{table}[!th]
\centering
\caption{Caption-generation performance on 25 test samples. Higher values indicate better performance.}
\label{tab:caption_results}
\begin{tabular}{lccccccc}
\midrule
Model & BLEU-1 $\uparrow$ & BLEU-2 $\uparrow$ & BLEU-3 $\uparrow$ & BLEU-4 $\uparrow$ & ROUGE-L $\uparrow$ & METEOR $\uparrow$ & CIDEr $\uparrow$ \\
\midrule
VideoChat2 & 0.221618 & 0.107851 & 0.061860 & 0.032972 & 0.160219 & 0.136542 & 0.004853 \\
VideoChat2 + GPEC & 0.317424 & 0.215904 & 0.152180 & 0.114343 & 0.253954 & 0.248644 & 0.008075 \\
\midrule
\end{tabular}
\end{table}

Across all reported caption-level metrics, VideoChat2 + GPEC achieves higher scores than the original VideoChat2 baseline. BLEU-1 increases from 0.2216 to 0.3174, while BLEU-2 increases from 0.1079 to 0.2159 and BLEU-3 increases from 0.0619 to 0.1522. BLEU-4 increases from 0.0330 to 0.1143. ROUGE-L increases from 0.1602 to 0.2540, while METEOR increases from 0.1365 to 0.2486. CIDEr also increases from 0.0049 to 0.0081. These results indicate improved agreement between the generated captions and the annotation-derived reference captions across multiple complementary caption-generation metrics.

The improvement is particularly pronounced for the higher-order BLEU metrics. BLEU-4 increases by more than threefold relative to the baseline, indicating greater overlap in longer $n$-gram sequences in addition to improvements in individual-word overlap. However, the different metrics capture complementary aspects of caption similarity and should therefore be interpreted jointly rather than through their absolute numerical scales.

\subsubsection{Content-Oriented Results}

Table~\ref{tab:content_results} reports the content-oriented evaluation over the same 25 test samples.

\begin{table}[!th]
\centering
\caption{Content-oriented evaluation of generated captions on 25 test samples. Higher values indicate better performance.}
\label{tab:content_results}
\begin{tabular}{lccc}
\midrule
Model & Field Recall $\uparrow$ & Exact Attribute Accuracy $\uparrow$ & Attribute Agreement $\uparrow$ \\
\midrule
VideoChat2 & 0.090909 & 0.000000 & 0.045455 \\
VideoChat2 + GPEC & 0.343434 & 0.257576 & 0.277778 \\
\midrule
\end{tabular}
\end{table}

The content-oriented evaluation shows that the effect of GPEC is not limited to lexical similarity. Clinical Field Recall increases from 0.0909 for VideoChat2 to 0.3434 for VideoChat2 + GPEC, indicating that a larger proportion of the predefined reference fields are detected in the corrected captions. Exact Clinical Attribute Accuracy increases from 0.0000 to 0.2576, showing that the corrected captions reproduce a larger proportion of the reference attributes exactly. Clinical Attribute Agreement also increases from 0.0455 to 0.2778, indicating that the predicted attributes more frequently correspond to the reference values under the predefined field-specific agreement rules, including cases in which the prediction is not an exact match but represents a closer qualitative level.

These results are consistent with the observed caption-level behavior of the model. The corrected representations lead to captions that more frequently contain the predefined cardiac content represented in the reference descriptions. At the same time, the lower attribute-level scores indicate that fine-grained qualitative characteristics remain more difficult to reproduce than broader content fields.

\subsubsection{Overall Quantitative Interpretation}

Taken together, the results show consistent improvement after introducing GPEC across all reported caption-level and content-oriented metrics. The improvements are observed in both lower- and higher-order lexical overlap measures, as well as in the recovery of predefined content fields and qualitative attributes.

The results support the use of pre-LLM embedding correction as a mechanism for improving the information available to the language-generation stage without modifying the original VideoChat2 backbone. The content-oriented results further indicate that the improvement is not restricted to surface-level caption similarity, while the remaining gap in exact attribute recovery suggests that representation correction does not fully recover all fine-grained qualitative information contained in the reference descriptions.

\subsection{Qualitative Results}

To complement the quantitative evaluation, six representative test samples are presented with their full generated captions. For each sample, the output of the original VideoChat2 model, the output obtained after applying GPEC, and the corresponding annotation-derived ground-truth caption are shown side by side. These examples are selected to illustrate the qualitative changes introduced by the pre-LLM embedding correction.

Across the selected examples, the baseline VideoChat2 captions frequently misidentify the imaging context or describe anatomical regions unrelated to the cardiac content of the input videos. For example, in several cases, the baseline refers to ultrasound examinations of the neck, thyroid, or other non-cardiac structures. In contrast, after applying GPEC, the generated captions consistently shift toward a cardiac interpretation, with references to the cardiac cycle, the left ventricle, diastole, and systole appearing in the generated descriptions.

The qualitative examples also illustrate that the correction is not limited to identifying the general cardiac context. In several cases, the GPEC-based captions additionally mention ventricular filling, changes in ventricular cavity size, contraction, and ventricular emptying. These descriptions are more closely aligned with the information represented in the corresponding reference captions. However, the generated captions do not consistently reproduce all fine-grained qualitative attributes specified by the references, such as the exact degree of filling, the magnitude of cavity reduction, or the precise level of systolic function and pumping performance.

Overall, the qualitative examples provide direct evidence of a consistent shift in the generated descriptions from broad or incorrect non-cardiac interpretations toward cardiac and left-ventricular content after pre-LLM embedding correction. At the same time, they demonstrate that GPEC improves the relevant content context without fully recovering every detailed attribute contained in the annotation-derived references.

The six representative examples are provided below.

\textbf{Example 1: 0X102CFB07F752AAE6:\\}
\textbf{VideoChat2:} \textit{[The video appears to be a grayscale ultrasound transverse view (Tv) of the neck, capturing the movement of the hyoid bone and surrounding structures. The image is likely taken during a medical examination, possibly to assess the function and position of the hyoid bone and its associated muscles and ligaments. The video duration is approximately 4 seconds, and it features a series of frames sampled at different intervals, ranging from 0]}

\textbf{VideoChat2 + GPEC:} \textit{[The video provided appears to be a detailed visual representation of the cardiac cycle, specifically focusing on the changes in the left ventricle's size and the flow of blood through the heart. The cardiac cycle consists of two main phases: diastole and systole. In the video, the left ventricle is shown in various stages of the cardiac cycle. During diastole, the left ventricle is relaxed, and blood flows]}

\textbf{Ground\_Truth:} \textit{[The video shows the cardiac motion of the heart throughout the cardiac cycle that should be discribed accurately. The left ventricle shows moderate filling during diastole, followed by a moderate reduction in ventricular cavity size during systole. The cavity becomes moderate after contraction, indicating moderately effective ventricular emptying. Overall, the observed ventricular motion is consistent with mildly reduced left ventricular systolic function and moderate pumping performance.]}

\textbf{Example 2: 0X10A28877E97DF540:\\}
\textbf{VideoChat2:} \textit{[The video appears to be a grayscale ultrasound (ultrasonography) image, likely taken during a medical examination. The process of obtaining this image involves the following steps: 1. The patient is positioned on an examination table, and a gel is applied to their skin to ensure proper contact between the ultrasound probe (transducer) and the body. 2. The ultrasound technician or physician places]}

\textbf{VideoChat2 + GPEC:} \textit{[The video appears to be a detailed visualization of the cardiac cycle, specifically focusing on the electrical activity of the heart and the corresponding mechanical events. The cardiac cycle consists of two main phases: diastole (relaxation and filling of the heart chambers) and systole (contraction and ejection of blood). The video aims to accurately depict these phases and the changes in the heart's electrical and mechanical activity.]}

\textbf{Ground\_Truth:} \textit{[The video shows the cardiac motion of the heart throughout the cardiac cycle that should be discribed accurately. The left ventricle shows substantial filling during diastole, followed by a marked reduction in ventricular cavity size during systole. The cavity becomes moderate after contraction, indicating effective ventricular emptying. Overall, the observed ventricular motion is consistent with preserved left ventricular systolic function and good pumping performance.s]}

\textbf{Example 3: 0X11DD5C8CC2A4D225:\\}
\textbf{VideoChat2:} \textit{[The video appears to be a grayscale ultrasound transverse view (TTE) of the heart, likely taken during a cardiac examination. The process of obtaining this video involves several steps: 1. Patient Preparation: The patient is positioned on an examination table, and the area of interest (the chest) is cleaned and draped to ensure a clear view of the heart. 2. Gel Application: A water-based gel is applied to the patient's chest to help transmit the ultrasound waves and reduce air gaps between the transducer and the skin. 3. Transducer Placement: A transducer, which is a
]}

\textbf{VideoChat2 + GPEC:} \textit{[The video you've provided appears to showcase the cardiac cycle, specifically focusing on the left ventricle's behavior during diastole and systole. The cardiac cycle is the sequence of events that occur when the heart beats, involving the contraction and relaxation of the atria and ventricles to pump blood throughout the body. Here's a detailed description of the process shown in the video: 1. Diastole (relaxation phase): The video begins by showing the left ventricle in a relaxed state during diastole. During this phase, the ventricle fills with blood from the left atrium through the mitral valve. The video shows]}

\textbf{Ground\_Truth:} \textit{[The video shows the cardiac motion of the heart throughout the cardiac cycle that should be discribed accurately. The left ventricle shows substantial filling during diastole, followed by a marked reduction in ventricular cavity size during systole. The cavity becomes moderate after contraction, indicating effective ventricular emptying. Overall, the observed ventricular motion is consistent with preserved left ventricular systolic function and good pumping performance.
]}\\

\textbf{Example 4: 0X105B9EF57DE45DCB:\\}
\textbf{VideoChat2:} \textit{[The video appears to be a grayscale ultrasound transverse view (T.V.) of the neck, capturing the movement of the thyroid gland. The video is 16 frames long, with a frame rate of 0.1 seconds per frame, resulting in a total duration of 1.6 seconds. The thyroid gland, located in the neck, is visible in the video, and it appears to be moving slightly. The video does]}

\textbf{VideoChat2 + GPEC:} \textit{[The video showcases the cardiac motion of the heart throughout the cardiac cycle, which should be displayed accurately. The left ventricle is depicted filling during diastole, followed by a decreased ventricular cavity size during systole. The cavity becomes smaller after contraction, which is a crucial part of the cardiac cycle. The video starts with the left ventricle in a relaxed state during diastole, where]}

\textbf{Ground\_Truth:} \textit{[The video shows the cardiac motion of the heart throughout the cardiac cycle that should be discribed accurately. The left ventricle shows relatively limited filling during diastole, followed by a moderate reduction in ventricular cavity size during systole. The cavity becomes moderate after contraction, indicating moderately effective ventricular emptying. Overall, the observed ventricular motion is consistent with mildly reduced left ventricular systolic function and moderate pumping performance.]}

\textbf{Example 5: 0X112EF236E1F676E4:\\}
\textbf{VideoChat2:} \textit{[The video appears to be a grayscale ultrasound (ultrasonography) image, likely taken using a transverse (T) mode. The process involves using high-frequency sound waves to create images of the internal structures of the body. The video shows a series of frames sampled at different time intervals, ranging from 0.1 to 2.7 seconds. These frames are likely used to create a moving image of the internal structures being examined]}

\textbf{VideoChat2 + GPEC:} \textit{[The video appears to be a detailed visualization of the cardiac cycle, specifically focusing on the left ventricle's behavior during systole and diastole. The cardiac cycle is the sequence of events that occur as the heart beats, involving the contraction (systole) and relaxation (diastole) of the atria and ventricles to pump blood throughout the body. In the video, the left ventric]}

\textbf{Ground\_Truth:} \textit{[The video shows the cardiac motion of the heart throughout the cardiac cycle that should be discribed accurately. The left ventricle shows moderate filling during diastole, followed by a marked reduction in ventricular cavity size during systole. The cavity becomes moderate after contraction, indicating effective ventricular emptying. Overall, the observed ventricular motion is consistent with preserved left ventricular systolic function and good pumping performance.]}

\textbf{Example 6: 0X126C02C8B984672B::\\}
\textbf{VideoChat2:} \textit{[The video appears to be a transverse view (T-V) ultrasound, a type of medical imaging technique used to visualize internal structures of the body. The video is presented in a grayscale format, with a time duration of 2.9 seconds. The frames are sampled at various intervals, ranging from 0.1 to 2.9 seconds. In this video, the ultrasound probe is likely placed on the]}

\textbf{VideoChat2 + GPEC:} \textit{[The video appears to be a visual representation of the cardiac cycle, specifically focusing on the left ventricle and its changes in size during diastole and systole. The cardiac cycle consists of two main phases: diastole, when the heart relaxes and fills with blood, and systole, when the heart contracts and pumps blood out. During diastole, the left ventricle is relaxed, and the mitral]}

\textbf{Ground\_Truth:} \textit{[The video shows the cardiac motion of the heart throughout the cardiac cycle that should be discribed accurately. The left ventricle shows substantial filling during diastole, followed by a marked reduction in ventricular cavity size during systole. The cavity becomes moderate after contraction, indicating effective ventricular emptying. Overall, the observed ventricular motion is consistent with preserved left ventricular systolic function and good pumping performance.]}

\subsection{Error Analysis}

\paragraph{Baseline errors.}

The original VideoChat2 outputs exhibit a recurring error pattern in which the anatomical or imaging context of the input video is incorrectly identified. In several representative samples, the baseline describes the videos as non-cardiac ultrasound examinations, including references to the neck, thyroid, or other unrelated anatomical structures. These errors indicate a substantial mismatch between the visual content of the cardiac videos and the semantic context expressed in the generated captions.

\paragraph{GPEC improvements and remaining errors.}

After introducing GPEC, the generated captions more consistently shift toward the cardiac context represented in the reference descriptions. In the qualitative examples, the corrected captions identify the cardiac cycle and more frequently refer to the left ventricle, diastole, systole, ventricular filling, cavity-size changes, and ventricular motion. However, the correction does not consistently recover all fine-grained attributes. For example, the generated captions may describe ventricular filling or cavity reduction without reproducing the exact qualitative level specified in the reference, such as `moderate,'' `marked,'' or `substantial.'' Similarly, functional descriptors such as `preserved,'' `mildly reduced,'' or `good'' may be omitted or simplified.

\paragraph{Interpretation and limitations.}

The observed behavior is consistent with the role of GPEC as a pre-LLM representation correction mechanism. The module modifies the projected visual representation before it is provided to the frozen language model, rather than explicitly injecting predefined clinical attributes into the generated text. Consequently, the correction can improve the semantic context available to the language-generation stage without guaranteeing complete recovery of every fine-grained attribute represented in the reference captions.

The content-oriented evaluation should also be interpreted within the design of the proposed metrics. These measures are based on a fixed ontology derived from the annotation-based reference captions and use rule-based lexical matching to detect predefined fields and attributes. They therefore provide a complementary analysis of content preservation rather than a general measure of semantic understanding or clinical reasoning. The observed improvements indicate better alignment with the predefined reference content, but they should not be interpreted as evidence of clinical diagnostic performance.

\section{Conclusion}
\label{conclusion}

In this work, we presented Gaussian Process Embedding Correction (GPEC), a lightweight and modular pre-LLM correction framework for improving caption generation from cardiac echocardiography videos using VideoChat2. GPEC operates on the projected visual representation immediately before the language model and learns a residual correction toward an annotation-guided target representation, while keeping the original VideoChat2 components frozen. The experimental results show consistent improvements after applying GPEC across representation-level, caption-level, and content-oriented evaluations. The corrected configuration achieves higher BLEU, ROUGE-L, METEOR, and CIDEr scores than the original VideoChat2 baseline, while the content-oriented analysis shows improved recovery of predefined cardiac fields and qualitative attributes. The qualitative examples further demonstrate a shift from incorrect non-cardiac interpretations toward descriptions that are more closely aligned with the cardiac content of the input videos. At the same time, the results show that fine-grained attribute recovery remains challenging. Corrected captions may identify the relevant cardiac context and ventricular motion while omitting or simplifying specific qualitative characteristics represented in the reference descriptions. This indicates that pre-LLM representation correction can improve broad content alignment without fully recovering all detailed information. Overall, GPEC demonstrates the potential of modular pre-LLM representation correction for specialized medical video caption generation without modifying the pretrained multimodal backbone. Future work can explore richer target representations, improved uncertainty modeling, more expressive kernel formulations, and broader evaluation across medical video domains and captioning tasks.

\bibliographystyle{unsrt}  
\bibliography{references}

\end{document}